\def\year{2018}\relax
\documentclass[letterpaper]{article} 
\usepackage{aaai18}  
\usepackage{times}  
\usepackage{helvet}  
\usepackage{courier}  
\usepackage{url}  
\usepackage{graphicx}  
\usepackage{algorithm}
\usepackage{algorithmicx}
\usepackage{algcompatible}
\usepackage{algpseudocode}
\usepackage{mathtools}
\usepackage{amsmath}
\usepackage[x11names,dvipsnames,table]{xcolor}
\usepackage{colortbl}
\usepackage{epsfig}
\usepackage{amssymb}
\usepackage{listings}
\usepackage{makecell}
\usepackage{booktabs}
\usepackage{multirow}
\usepackage{color}
\usepackage{comment}
\usepackage{icomma}
\usepackage{enumitem}
\usepackage{float}

\begin{document}
	
	\title{Kill Two Birds with One Stone: Weakly-Supervised Neural Network \\for Image Annotation and Tag Refinement}
	\author{Junjie Zhang$^{1,2}$ \quad Qi Wu$^{3}$ \quad Jian Zhang$^{1}$ \quad Chunhua Shen$^{3}$ \quad Jianfeng Lu$^{2}$\\
		$^1$Faculty of Engineering and Information Technology, University of Technology Sydney, Australia \quad \\ 
		$^2$School of Computer Science and Technology, Nanjing University of Science and Technology, China\quad \\
		$^3$Australian Centre for Robotic Vision, University of Adelaide, Australia \\
		{ \{junjie.zhang@student., jian.zhang@\}uts.edu.au  \quad \{qi.wu01, chunhua.shen\}@adelaide.edu.au} \quad
		{lujf@njust.edu.cn}
	}
	\maketitle

	\begin{abstract}
		The number of social images has exploded by the wide adoption of social networks, and people like to share their comments about them. These comments can be a description of the image, or some objects, attributes, scenes in it, which are normally used as the user-provided tags. However, it is well-known that user-provided tags are incomplete and imprecise to some extent. Directly using them can damage the performance of related applications, such as the image annotation and retrieval. In this paper, we propose to learn an image annotation model and refine the user-provided tags simultaneously in a weakly-supervised manner. The deep neural network is utilized as the image feature learning and backbone annotation model, while visual consistency, semantic dependency, and user-error sparsity are introduced as the constraints at the batch level to alleviate the tag noise. Therefore, our model is highly flexible and stable to handle large-scale image sets. Experimental results on two benchmark datasets indicate that our proposed model achieves the best performance compared to the state-of-the-art methods.
	\end{abstract}
	
	\section{Introduction}
	As the imaging technology tends to be perfect and the wide usage of social networks, a large number of images are shared through the Internet every day, including the landscape photos, selfies, snapshots and so on. However, a significant amount of them is unlabeled or weakly labeled.
	To better understand and efficiently retrieval these images, it is essential to develop an automatic annotation method. Traditional methods on the image annotation focus on using human-labeled images as training data to uncover the relationships between image visual content and tags \cite{guillaumin2009tagprop,makadia2010baselines}. In recent years, the deep neural network \cite{krizhevsky2012imagenet,simonyan2014very} has achieved superior performance on image feature learning and been widely used in the image classification and related vision tasks \cite{single2multi,wang2016cnn}. However, these deep models are purely based on the supervised learning and require a significant amount of well-labeled training samples to obtain satisfactory results. The labeling process can be intensive and expensive. It is unrealistic for the human to continuously label the large-scale images that pop into the social networks every day, to obtain high-quality training samples.
	
	\begin{figure}[t]
		\centering
		\includegraphics[width=1\linewidth]{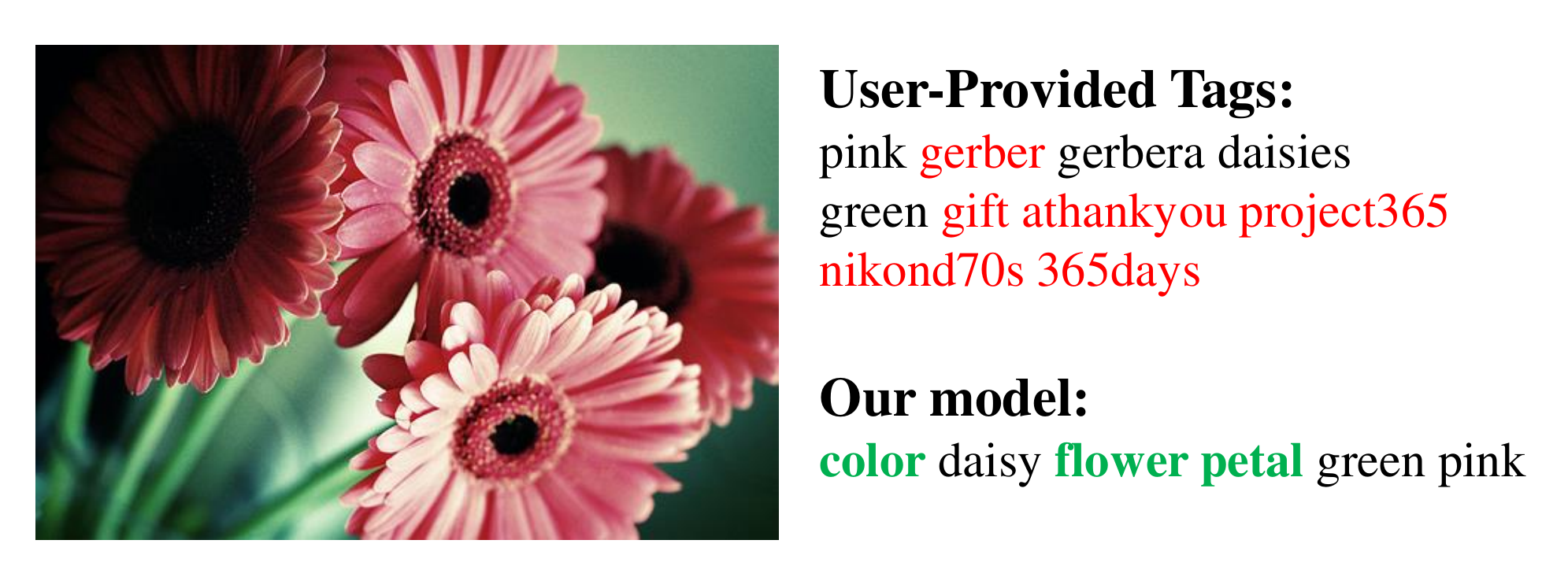}
		\caption{An example image from the Mirflickr dataset. Compare to our model's predictions, the user-provided tags are inaccurate, incomplete and some are meaningless for visual understanding, such as `gerber,' `project365', and `nikond70s'.
		}
		\label{example_img}
	\end{figure}
	
	\begin{figure*}[t]
		\centering
		\includegraphics[width=0.97\linewidth]{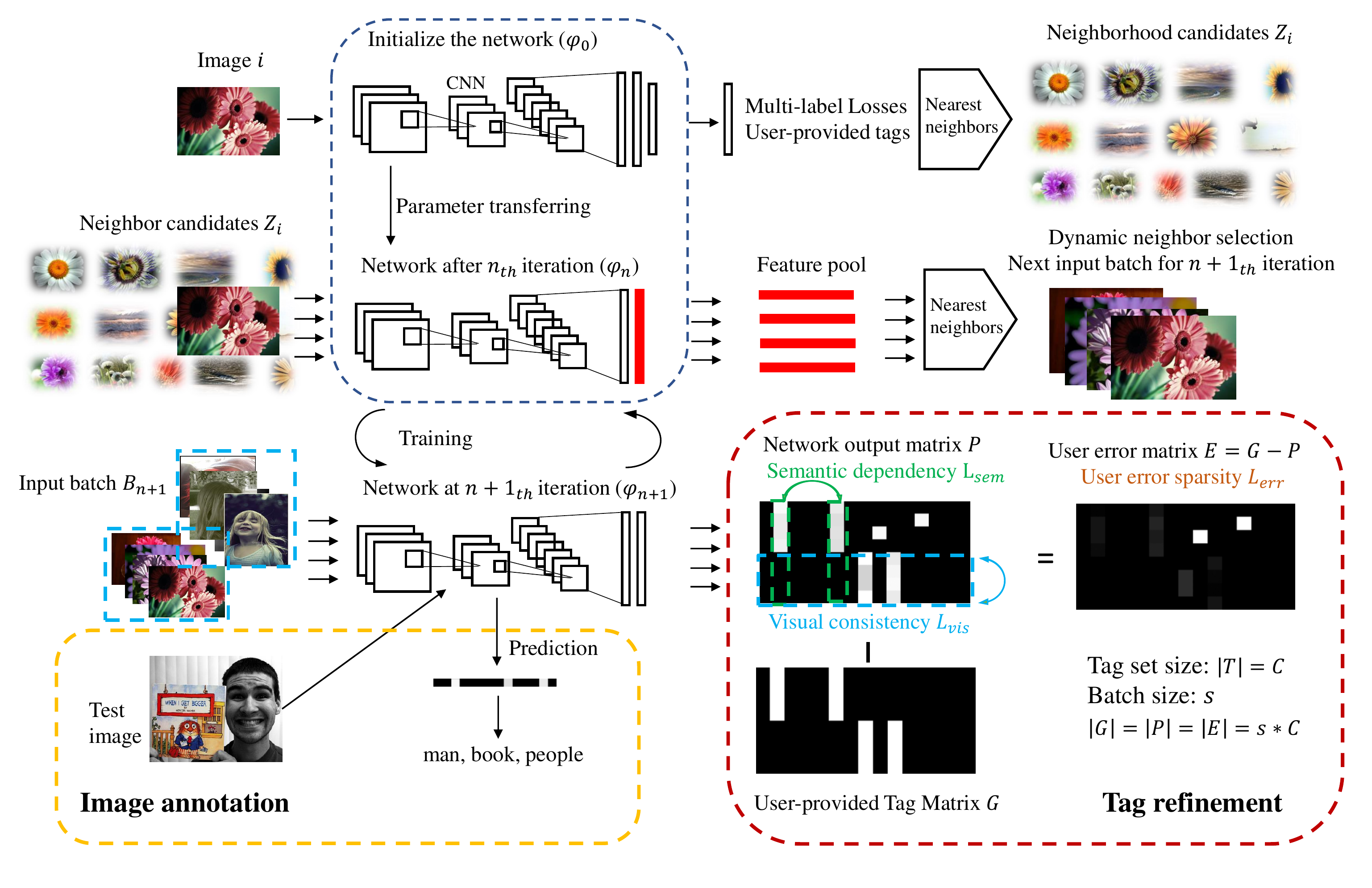}
		\caption{Our proposed weakly-supervised model. We adopt Deep Convolutional Neural Network as the backbone annotation model. The network $\varphi_0$ is initialized by training on the user-provided tags, and further transferred by training with the proposed constraints ($L_{vis}$, $L_{sem}$ and $L_{err}$). We dynamically choose the image neighbors from the neighborhood candidates and combine them as the input batch $B_{n+1}$ for network $\varphi_{n+1}$. The tag refinement is conducted during training, while the new image can obtain annotations by passing through the trained network.} 
		\label{framework}
	\end{figure*}

	Sometimes, people spontaneously assign tags to some images when uploading them, and more tags are attached along with the spread. Despite the fact that these tags provide semantic illustrations of the images to some extent, they are always incomplete, imprecise, and biased toward the individual perspectives \cite{li2016socializing}. See Fig.\ref{example_img} as an example. To alleviate the existing tag noise for further retrieval and training annotation model, some previous works are \cite{zhu2012sampling,liu2009tag,li2009learning,li2013classifying,zhu2010image} conducted on the tag relevance analysis and refinement. These works explore the tag relevance from different perspectives including the semantic similarities between tags, the visual similarities among image neighbors and the properties of the tagging matrix. The user-provided tag is further refined based on its relevance. The most relevant works are conducted in \cite{zhu2010image,li2017weakly,li2017weaklymatrix}. However, they intend to focus on the low-rank property of the tagging matrix, which neither can connect the image feature to the refined results nor enable the image feature learning. Therefore, these methods are less flexible and stable when facing the large-scale image sets.
	
	To address the above issues, in this paper, we propose to learn an image annotation model and refine the user-provided tags simultaneously in a weakly-supervised manner. The whole framework is shown in Fig.\ref{framework}. Deep Convolutional Neural Network (DCNN) is adopted as the feature learning and backbone annotation model. Different from the regular DCNN that is trained on the supervised information, which is usually professionally annotated and double checked, we consider the user-provided tags as weakly-supervised information to assist the training. We propose to learn the image visual representation and the relationship between visual content and tags by exploring the visual consistency among the image neighbors and the semantic dependencies between the tag pairs. That is, images with similar visual appearance are usually annotated with the same tags, and semantic dependent tags intend to appear in the same images jointly. To efficiently utilize these constraints and enable the feature learning, we propose to dynamically generate the neighbors for each image and form them as the input batches. Given the input batches, we apply these two constraints on the tag probability distributions generated by the neural network. Moreover, although the user-provided tags are noisy and biased, people share the general knowledge about the semantic annotation. The user-provided tags are still accurate at a reasonable level, and each image usually is assigned with very few tags compare to the entire tag set. Therefore, the error of a batch user-provided tags is sparse. By setting these constraints at the batch level, we can train the neural network to conduct tag refinement and learn the annotation model at the same time. In summary, the main contributions of our model are as follows:
	
	1) We propose to obtain the deep neural network based image annotation model and conduct the tag refinement simultaneously in a weakly-supervised manner. During the training, the user-provided tags are spontaneously refined to probability distributions, while the trained annotation model can be applied to assign tags to new images.
	
	2) We set the constraints of the neural network at the batch level, which not only enable the image feature learning but also make the model flexible and stable when handle the large scale user-annotated training samples at a low computation cost.
	
	3) Our proposed model achieves the state-of-the-art performance for both image annotation and tag refinement experiments on two benchmark datasets.
	
	\section{Related Works}
	
	High-quality tags of images are necessary for the image understanding and retrieval. Various works have been conducted on analyzing the tag relevance, improving the tag quality and automatically annotating images. Early works intend to estimate the tag relevance based on the semantic information only. In \cite{zhu2012sampling}, tag relevance is evaluated by averaging the WordNet \cite{miller1995wordnet} similarities between the assigned tags for each image, while latent Dirichlet allocation model \cite{blei2003latent} and collective knowledge are used in \cite{xu2009tag} and \cite{sigurbjornsson2008flickr} respectively. However, these methods overlook the image visual information and highly rely on the initial tags of images, which are limited to annotating new images.
	
	Many approaches have been proposed by leveraging the visual information along with the associated tags. In \cite{liu2009tag}, the initial probabilistic tag relevance is estimated by kernel density estimation; the random walk is performed based on both visual and semantic similarities to rank the tags in favor of the retrieval. In \cite{makadia2010baselines}, the nearest neighbor voting mechanism is employed to assign the tags to new images based on the visual similarity, while in \cite{guillaumin2009tagprop}, instead of treating neighbors equally, distance metric learning is used to reweight them. Li et al. \cite{li2013classifying} propose to select relevant positive training samples and negative samples from the noisy tags to train the classifier for annotation. Positive samples are selected based on the aforementioned voting methods, while the negative ones are collected by bootstrap.
	
	There are also some works that focus on the modality design. For example, in \cite{chen2012tag}, the image feature is enriched by adding additional tag feature, which is obtained by the SVM prediction, while in \cite{pereira2014role} and \cite{ballan2014cross}, authors design a latent multi-modal space to tackle the annotation problem by canonical correlation analysis and kernel canonical correlation analysis. Besides, matrix-based methods are also widely used for the tag refinement. Image tag as one type of semantic information is subject to the low-rank property. Therefore, the initial tagging matrix can be decomposed into the ideal tagging matrix with a sparse user-error matrix \cite{zhu2010image,li2017weaklymatrix}. In \cite{zhu2010image}, they use the predefined visual and semantic similarity to assist the process. However, these methods fail to connect the visual features with the tagging results, which makes it unable to perform the annotation. In \cite{li2017weakly}, a three-layer network architecture is proposed to bridge the semantic gap. However, since it cannot perform feature learning and use the matrix as the input, it is less flexible and stable when dealing with the large-scale image sets.
	
	Inspired by the advanced abilities of the deep neural network, various models have been proposed for vision and multimedia tasks, especially for image annotation and retrieval, such as \cite{gong2013deep} and \cite{wan2014deep}. However, these deep models rely on the high-quality tags as supervised information, which is hard and expensive to obtain. Different from the above methods, in our work, we use the deep neural network as the feature learning and backbone annotation model, while dynamically constrain the network at the batch level in a weakly-supervised manner.
	
	\section{Proposed Model}
	
	\subsection{Overview} 
	The key characteristic of our model is that we formulate constraints of the deep neural network from two aspects. One is the internal relationships of the image set, which is reflected as the visual consistency among image neighbors and the semantic dependencies between tag pairs. The other is the general knowledge that error of the user-provided tags is sparse. To appropriately introduce these constraints into the neural network, we choose the input batches dynamically to enable the image feature learning. The entire model is shown in Fig.\ref{framework}. 
	
	\subsection{Initialization of the Network}
	We first train the network as a regular multi-label neural network on the user-provided tags. The motivations for doing this are twofold. First, we want to give a relatively good initialization of the network parameters, since the deep neural network can achieve superior performance on feature learning and annotation owe to it is composed of multiple nonlinear transformations with a huge number of parameters. Second, we want to find the neighborhood candidates for each image using obtained visual features, which is a necessary step in the visual consistency part.
	
	Let $I$ be the image set, $T$ be the set of possible initial tags provided by the users, and $D = \{(i,t) | i \in I, t \in T \}$ be the image dataset associated with these user-provided tags, where $|I| = N$ and $|T| = C$, corresponding to the image and tag set size respectively. For each image $i$ with the user-provided ground-truth vector $y_i = [y_{i1}, y_{i2}, \dots, y_{iC}]$ ($y_{ij} = 1$ if image is annotated by the $j_{th}$ tag, otherwise it is $0$), we use CNN ($\varphi$) to extract image visual feature followed by a fully-connected layer with a sigmoid transformation to generate a $C$-dimensional vector to represent the tag probability distribution. The logistic loss is employed to train the network. We note this initial model as $\varphi_0$.
	
	After the initialization is finished, we start to train the neural network by considering the visual consistency $L_{vis}$ with semantic dependency $L_{sem}$ and user-error sparsity $L_{err}$  altogether; we give the final form of our network constraints:
	\begin{equation}
	L_{final} = L_{vis} + \lambda_1 L_{sem} + \lambda_2 L_{err}
	\end{equation}
	where $\lambda_1$ and $\lambda_2$ are set to balance the different constraints. In the following subsections, we first introduce the dynamic selection of neighbors and the visual consistency. The semantic dependency is presented next, and followed by the user-error sparsity constraint. Finally, we summarize how the proposed model performs the tag refinement and predict tags for new images. Implementation details will also be given in this section.
	
	\subsection{Dynamic Neighbors and Visual Consistency}
	\subsubsection{Dynamic Neighbors}
	We generate image visual neighbors and input batches by dynamically using the nearest neighbors approach. Given the initialized neural network $\varphi_0$, we first extract $d$-dimensional visual features of the whole image set $I$ as $V = \{ v_i | v_i = \varphi_0(i), i\in I \}$. We use the Euclidean distance between each feature pair to rank the initial neighborhood candidates. 
	
	Let $Z_{i}$ be the initial candidate set of image $i$. Since the neural network learns to extract image visual features in our model, the parameters of the network are updated after each iteration, which means the visual feature of each image is also changed, so as the neighbors. Because we intend to constrain the network among neighbors, it would be time-consuming to forward the whole training set and perform the nearest neighbors approach after each iteration. To efficiently apply the training process, after each iteration, we forward the initial candidate set $Z_i$ of next input image $i$ to update the neighbors; we set $|Z_i|=M (M<N)$; then we form the next input batch. The input generation process is described in Algorithm.\ref{input generation}.
	
	\begin{algorithm}[t] 
		\caption{Generate Input Batch for Network Training}
		\label{input generation} 
		\begin{algorithmic}
			\Require\\
			$\varphi_{n}:$ Neural network after $n_{th}$ iteration;\\
			$s:$ Batch size;\\
			$m:$ Size of the final neighbors, $m<M$;\\
			${i_1, i_2, \dots, i_{\frac{s}{m}}}:$ Next input images;\\
			$Z_{i_{j}}:$ Neighborhood candidates of $j_{th}$ image, $j\in[1,\frac{s}{m}]$;\\
			\Ensure
			$B_{n+1}:$ Next input batch for $n+1_{th}$ iteration;\\
			\For{Each $j\in[1,\frac{s}{m}]$} 
			\State 1: Forward candidates $Z_{i_{j}}$ to update image features as $V'_{j} = \{  v'_{i_j} | v'_{i_j} = \varphi_{n}(i_{j}) \}$; 
			\State 2: Nearest neighbors approach in $V'_{j}$ to update $Z_{i_{j}}$;
			\State 3: Select top $m$ images as mini-batch $B_{n+1,j}$;
			\EndFor
			\State Concatenate all the mini-batches as input batch $B_{n+1}$.
		\end{algorithmic}
	\end{algorithm}
	
	\subsubsection{Visual Consistency}
	After generating the input batches of the neural network, now we introduce the first constraint of our model. Based on the observation that visual similar images intend to be annotated with the similar tags (i.e., tag distributions should be close), we define the visual consistency constraint as follows.
	
	Each mini-batch $B_{n+1,j}$ is composed of the image $i_j$ with its neighbors $Z_{i_j} = \{ z_{e} | e \in [1,\frac{s}{m}] \}$, where $s$ is the batch size, and $m$ is the number of the image $i_j$ plus its selected neighbors. The visual similarity between image $i_j$ and $z_e$ is defined as $\gamma_{i_j,e} = exp(-\frac{\lVert \varphi(i_j)-\varphi(z_e) \rVert^2}{\sigma})$, $\sigma$ is the medium value of $\gamma$. Then the visual consistency constraint can be carried out as:
	\begin{equation}
	L_{vis} = \mathop {\min }\limits_P \frac{m}{s(m-1)} \sum_{j=1}^{s/m} \sum_{e=1}^{m-1} \gamma_{i_j,e} \lVert p_{i_j}-p_{z_e} \rVert^2
	\end{equation}
	where $p_{i_j}$ stands for the tag probability prediction of the image $i_j$, $|p_{i_j}|=C$. $P$ is the tag probability prediction of the whole batch, $|P|=s*C$.
	
	\subsection{Semantic Dependency}
	Besides the visual consistency among images, we also consider the semantic dependencies of the tag pairs. It is natural that social tags are not assigned separately, semantically similar tags often appear together in the similar images. Based on this knowledge, we first estimate the tag-pair similarity.
	
	We consider the tag-pair similarity from two aspects: context and knowledge base. Given two tags $t_i$ and $t_j$, context ($dist_{ctx}$) is defined as the Google distance \cite{cilibrasi2007google} of two tags in the given set (we use the image instead of the web page), while the knowledge base ($dist_{KB}$) is the WordNet similarity \cite{miller1995wordnet} based on the information content of the least common subsumer and input synsets. The reason we use two similarity metrics is to transfer the general measurement to the collected set. That is:
	
	\begin{align}
	&dist_{ctx}(t_i, t_j)= \frac{max(logf(t_i),logf(t_j))-logf(t_i,t_j)}{logN-min(logf(t_i),logf(t_j))}\\
	&dist(t_i, t_j)=dist_{ctx}(t_i, t_j)+ \alpha dist_{KB}(t_i, t_j)\\
	&\xi_{t_{ij}}=exp ( -dist(t_i, t_j)^2/\sigma )
	\end{align}
	
	where $f(t_i)$ is the frequence of tag $t_i$ in dataset $D$, $f(t_i,t_j)$ is the co-occurrence of tag pair $(t_i, t_j)$, $\alpha$ is set to balance two metrics, $\sigma$ is the medium value of $\xi$. It is worth noting that $\xi$ is a symmetrical matrix, which is pre-computed as a look-up table for training. Then the semantic dependency constraint on the input batch is carried out as:
	
	\begin{equation}
	L_{sem} = \mathop {\min }\limits_P \frac{1}{C^2} \sum_{i=1}^{C} \sum_{j=1}^{C} \xi_{t_{ij}} \lVert p'_{t_i}-p'_{t_j} \rVert^2
	\end{equation}
	
	where $p'_{t_i}$ stands for the probabilities of the image batch annotated by $t_i$, $|p'_{t_i}|=s$. To speed up the training process, by referring to \cite{zhu2010image}, we use the matrix form of this constraint, let $Q$ be the diagonal matrix, then the semantic dependency constraint can be written as:
	
	\begin{align}
	&Q_{ii} = \sum_{i\neq j}\xi_{t_{ij}} \quad i,j \in [1, C]\\
	&L_{sem} = \mathop {\min }\limits_P \frac{1}{C^2} Tr[P^T(Q-\xi)P]
	\end{align}
	
	\subsection{User-Error Sparsity}
	Although the user-provided tags are relatively noisy and biased, people share the general knowledge about semantic annotation. The user-provided tags are still accurate at a reasonable level. Moreover, each image is usually assigned with few tags compare to the entire tag set. Therefore, the error of a batch user-provided tags is sparse. Let $G(|G|=s*C)$ be the user-provided annotation matrix of the input batch, which has the same dimensions as the network probability output $P$. Each row vector in $G$ represents the user annotation for each image. The difference matrix between $G$ and $P$ is the user-error matrix. Thus, the sparsity constraint is defined as follows:
	
	\begin{equation}
	L_{err} = \mathop {\min }\limits_P \lVert G-P \rVert_1
	\end{equation}
	
	\subsection{Training and Prediction}
	We use VGG-16 as our backbone neural network $\varphi$ to conduct image feature learning. The training process is two-staged: first, we obtain $\varphi_0$, the output of last fully-connected layer $(4096-d)$ is used as image feature to perform the nearest neighbor approach to find the neighborhood candidates, $\varphi_0$ is also used to initialize the network for weakly-supervised training. Then we train $\varphi$ with the proposed constraints and dynamically generate the input batches. The whole training process is shown in Algorithm.\ref{Training Process}.
	
	\begin{algorithm}[H] 
		\caption{Network Training Process}
		\label{Training Process} 
		\begin{algorithmic}[1]
			\State Train $\varphi_0$ with the multi-label logistic loss;
			\State Compute the initial neighborhood candidates $Z_{i_j}$ for each image $i_j$;
			\State Initialize $\varphi$ with $\varphi_0$;
			\State Dynamically generate input batch $B_{n+1}$ as Algorithm.\ref{input generation};
			\State Train $\varphi_{n+1}$ with $L_{final}$;
		\end{algorithmic}
	\end{algorithm}
	
	We train all the models for thirty epochs, with learning rate 0.001 from the start and decrease it to one-tenth every ten epochs. Stochastic gradient descent \cite{bottou2010large} is used to optimize the models. The grid-search strategy is adopted to tune the hyperparameters including $\lambda_1, \lambda_2$ and $\alpha$ by referring to the previous works \cite{zhu2010image,li2017weakly}. Tag refinement proceeds naturally during the $\varphi$ training. For the image annotation, new images are sent into the trained network to obtain the tag probability distributions. Moreover, we visualize the proposed constraints based on the experimental results. The details are introduced in the supplementary materials.
	
	\section{Experiments}
	In this section, we present our experimental results and analyze the effectiveness of our proposed model. Our model is evaluated on two benchmark datasets: Mirflickr \cite{huiskes2008mir} and NUS-WIDE \cite{chua2009nus}. By comparing with the baselines and the state-of-the-art models, we show that proposed model achieves the best performance.
	\subsection{Data Preprocessing}
	Images and social tags of the Mirflickr and NUS-WIDE dataset are obtained from the Flickr website. The tags are free-form and need to be unified to conduct adequate research. Besides, for a tag to be meaningful, it needs to be assigned to a certain number of images. Therefore, we carry out the preprocessing as follows to obtain training set: first, we lemmatize all the tags to their dictionary forms and remove the ones that do not appear in the WordNet, then we exclude the tags that do not meet the occurrence threshold ($0.1\%$ of total image number). We evaluate all the models on tags which are manually corrected (note as the label in this section). We choose one-fifth data for training and the rest for test. That is 5000 and 50,000 training images for Mirflickr and NUS-WIDE respectively. The experiments are repeated five times, and average results are reported. The statistics of two datasets are shown in Tab.\ref{dataset}:
	
	\begin{table}[t!]
		\centering
		\resizebox{\linewidth}{!}{
			\begin{tabular}{c c c c c}
				\toprule
				& Img number $N$  & Tag set size $C$ & Label number & Tags per img \\ 
				\midrule
				Mirflickr   & 25,000 & 444 & 14 & 2.7 \\
				\midrule
				NUS-WIDE   & 201,302 & 3010 & 81 & 6.8 \\
				\bottomrule
		\end{tabular}}
		\caption{The statistics of the Mirflcikr and NUS-WIDE after preprocessing, including the total image amount, the size of tag set and labels, and the number of tags per image.}
		\label{dataset}
	\end{table}

	\subsection{Evaluation Metrics}
	Several metrics are employed to evaluate the performance of the proposed model and the state-of-the-art methods. Results of the image annotation and the tag refinement are both reported. We refer to the previous works \cite{li2016socializing,li2017weakly} to compute the average precision (AP) and the area under the receiver operating characteristic curve (AUC).
	For each image, a good model should rank relevant tags before the irrelevant ones. Moreover, for a given tag query, relevant images should be returned first before the irrelevant ones. Therefore, we use the mean image average precision (miAP) and mean average precision (mAP) to measure the model performance. miAP is computed by averaging the APs on all the images, while mAP is computed by averaging the APs on all the given tags. Similar to AP, global and average performance of AUC is measured. MicroAUC is computed by concatenating all the tag probability vectors together and average the AUC, while MacroAUC is calculated by averaging the mean AUC of each given tag.
	
	\subsection{Baselines and Compared Methods}
	We give the descriptions of the baselines first and then list all the compared state-of-the-art methods of the image annotation and tag refinement:
	
	\textbf{RandomGuess:}
	\cite{li2016socializing} This is a baseline for the image annotation. Given a new image, RandomGuess assigns tags by randomly selecting from the tag set. We run RandomGuess eighty times, and evaluate it by averaging the predicted scores. 
	
	\textbf{UserTags:}
	\cite{li2016socializing} This is a baseline for the tag refinement. All the user-provided tags are reserved, and the performance is evaluated based on them. 
	
	\textbf{KNN:}
	\cite{makadia2010baselines} This is a baseline for the image annotation. KNN model measures the tag relevance respect to the given image by retrieving the $k$ nearest neighbors from the image set. Then the tags are assigned based on their occurrence rates among the neighbors. The image feature used in this model is the 4096-d vector extracted by VGG-16. 
	
	\textbf{Multi-CNN:} This baseline has the same configurations as our initialized model $\varphi_0$. For the image annotation, we train the neural network using one-fifth data with logistic loss and test it on the rest of set. For the tag refinement, the model is trained on the whole dataset to better evaluate the refinement performance for large-scale datasets. 
	
	\textbf{Compared Methods:} For the image annotation, we compare with several state-of-the-art methods, including TagProp \cite{guillaumin2009tagprop}, CCA \cite{murthy2015automatic}, TagFeature \cite{chen2012tag}, TagExample \cite{li2013classifying} and WDNL \cite{li2017weakly}. For the tag refinement, we compare with TagCooccur \cite{sigurbjornsson2008flickr}, TagVote \cite{li2009learning}, and RPCA \cite{zhu2010image}. For a fair comparison, the compared models and baselines are using the pre-trained VGG-16. We implement an equivalent model of WDNL on NUS-WIDE dataset named NMF by using fixed VGG-16 feature matrix as input with low rank matrix decomposition. 
	
	\subsection{Results on Image Annotation}
	For the image annotation, we train our model on one-fifth data as described in the last section. We set the batch size $s=64$, the initial neighborhood candidates size $M=512$ and the final neighbor size $m=8$.
	\begin{table}[h]
		\centering
		\resizebox{\linewidth}{!}{
			\begin{tabular}{ccccc}
				\Xhline{2\arrayrulewidth}
				Method & imAP & mAP & MicroAUC  & MacroAUC\\ \Xhline{2\arrayrulewidth}
				\midrule
				RandomGuess & 0.072 & 0.072 & 0.501 & 0.498\\
				KNN & 0.243 & 0.499 & 0.785 & 0.926\\
				Multi-CNN & 0.404 & 0.556 & 0.865 & 0.925\\\Xhline{2\arrayrulewidth}
				CCA & - &0.293 & 0.642 & 0.627\\
				WDNL & - & 0.382 & 0.665 & 0.652\\
				TagProp & 0.386 & 0.518 & 0.822 & 0.907\\
				TagFeature & 0.313 & 0.414 & 0.786 & 0.892\\
				TagExample & 0.324 & 0.537 & 0.728 & 0.915\\
				Our model & \textbf{0.449} & \textbf{0.591} & \textbf{0.893} & \textbf{0.941}\\
				\Xhline{2\arrayrulewidth}
		\end{tabular}}
		\caption{Image annotation results on the Mirflickr.}
		\label{mirflickr_annotation_res}
	\end{table}

	\begin{table}[h]
		\centering
		\resizebox{\linewidth}{!}{
			\begin{tabular}{ccccc}
				\Xhline{2\arrayrulewidth}
				Method & imAP & mAP & MicroAUC  & MacroAUC\\ \Xhline{2\arrayrulewidth}
				\midrule
				RandomGuess & 0.023 & 0.023 & 0.500 & 0.504\\
				KNN & 0.388 & 0.357 & 0.916 & 0.941\\
				Multi-CNN & 0.405 & 0.369 & \textbf{0.922} & 0.934\\\Xhline{2\arrayrulewidth}
				CCA & 0.363 &0.364 & 0.865 & 0.928\\
				NMF & 0.383 & 0.369 & 0.910 & 0.923\\
				TagProp & 0.359 & 0.373 & 0.921 & 0.930\\
				TagFeature & 0.240 & 0.302 & 0.831 & 0.906\\
				TagExample & 0.356 & 0.335 & 0.919 & 0.924\\
				Our model & \textbf{0.412} & \textbf{0.398} & \textbf{0.922} & \textbf{0.942}\\
				\Xhline{2\arrayrulewidth}
		\end{tabular}}
		\caption{Image annotation results on the NUS-WIDE.}
		\label{nus_annotation_res}
	\end{table}

	Tab. \ref{mirflickr_annotation_res} and \ref{nus_annotation_res} show that the proposed model achieves the best performance on all the evaluation metrics of both datasets. As we can see, all methods outperform the baseline RandomGuess, which proves that learning from the user-provided tags is useful for the image annotation. KNN model only uses the visual similarities among images, while Multi-CNN learns the classifier for each tag independently. Therefore, our model surpasses the KNN and Multi-CNN, which indicates that it is significant to explore the visual and semantic relationships inside the image set at the same time. Results against CCA shows that eliminate the tag noise is necessary for social image annotation. Prior works including the TagProp, TagFeature, and TagExample are proposed to perform annotation using the user-provided tags. However, our model outperforms these methods since we combine the deep network architecture with the weakly-supervised constraints. Different from the WDNL and NMF, which uses low-rank matrix decomposition with a three-layer neural network for the tag prediction, our model enables the neural network for image feature learning in an end-to-end fashion by formulating the constraints at the batch level, which achieves the significant better results. 

	\begin{figure*}[t]
		\centering
		\includegraphics[width=0.9\linewidth]{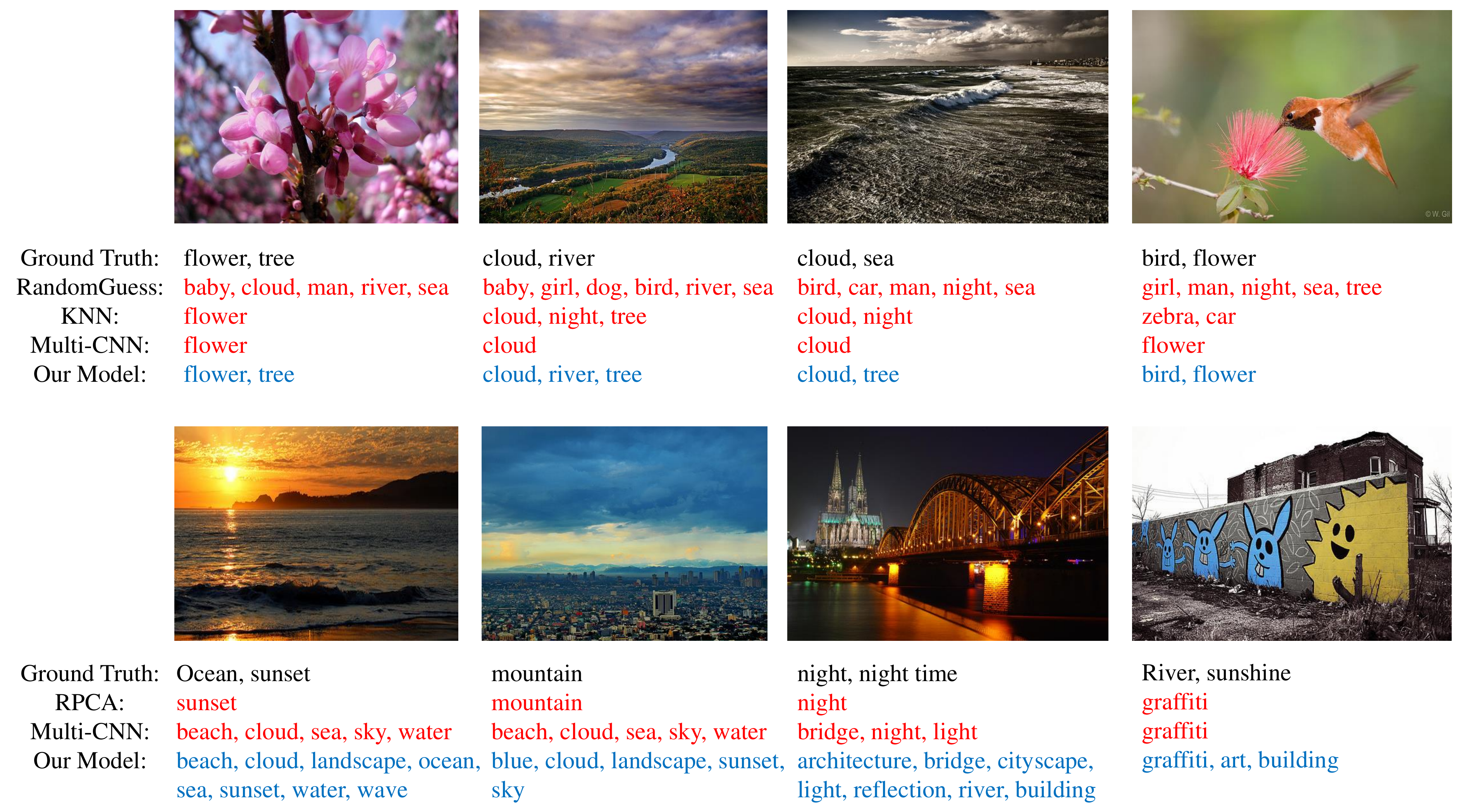}
		\caption{(a) First row: image annotation examples from the Mirflickr 14 labels. (b) Second row: tag refinement examples.}
		\label{results_examples} 
	\end{figure*}

	\subsection{Results on Tag Refinement}
	For the tag refinement, we train our model on the entire dataset to show the effectiveness of our model when refining the large-scale datasets. We set parameters same as the annotation experiment.
	
	\begin{table}[t]
		\centering
		\resizebox{\linewidth}{!}{
			\begin{tabular}{ccccc}
				\Xhline{2\arrayrulewidth}
				Method & imAP & mAP & MicroAUC  & MacroAUC\\ \Xhline{2\arrayrulewidth}
				\midrule
				UserTags & 0.100 & 0.263 & 0.544 & 0.642\\
				Multi-CNN & 0.426 & 0.597 & 0.872 & 0.937\\ \Xhline{2\arrayrulewidth}
				TagCooccur & 0.159 & 0.260 & 0.587 & 0.699\\
				TagVote & 0.201 & 0.323 & 0.594 & 0.708\\
				RPCA & 0.384 & 0.541 & 0.840 & 0.914\\
				Our model & \textbf{0.476} & \textbf{0.633} & \textbf{0.907} & \textbf{0.952}\\
				\Xhline{2\arrayrulewidth}
		\end{tabular}}
		\caption{Tag refinement results on the Mirflickr.}
		\label{mirflickr_refine_res}
	\end{table}

	\begin{table}[t]
		\centering
		\resizebox{\linewidth}{!}{
			\begin{tabular}{ccccc}
				\Xhline{2\arrayrulewidth}
				Method & imAP & mAP & MicroAUC  & MacroAUC\\ \Xhline{2\arrayrulewidth}
				\midrule
				UserTags & 0.187 & 0.338 & 0.656 & 0.783\\
				Multi-CNN & 0.424 & 0.416 & 0.921 & 0.935\\ \Xhline{2\arrayrulewidth}
				TagCooccur & 0.277 & 0.298 & 0.650 & 0.818\\
				TagVote & 0.311 & 0.368 & 0.905 & 0.864\\
				RPCA & 0.404 & 0.426 & 0.918 & 0.872\\
				Our model & \textbf{0.431} & \textbf{0.446} & \textbf{0.927} & \textbf{0.950}\\
				\Xhline{2\arrayrulewidth}
		\end{tabular}}
		\caption{Tag refinement results on the NUS-WIDE.}
		\label{nus_refine_res}
	\end{table}

	Tab. \ref{mirflickr_refine_res} and Tab. \ref{nus_refine_res} indicate that the proposed model outperforms the baselines and state-of-the-art models on both datasets. As we can see from two tables, the baseline UserTags directly uses user-provided tags without any refinement, while different refinement methods have shown different degrees of improvements against it. The TagCoocur only uses the tag co-occurrence rate and frequency to rank the tags respective to the image, no visual information involved, while the TagVote also considers the visual similarities among images to refine the ranking results. However, by dynamically choose the input neighbors, our model can perform the feature learning during the refinement, which outperforms these methods. The better results against the Multi-CNN proves the effectiveness of the proposed constraints. Moreover, we achieve better results compared to the most relevant refinement method RPCA. As illustrated in \cite{li2016socializing}, since the RPCA optimizes the whole tagging matrix, it could not be easily applied to the large-scale datasets due to its high demand in both CPU time and memory. On the contrary, we formulate three constraints at the batch level, which actives the feature learning and also make our model flexible and stable to deal with the large-scale datasets. We give some examples in Fig. \ref{results_examples} to show the refinement results.
	As we can see, our proposed model removes the inaccurate user-provided tags and adds relevant tags to images.
	
	\subsection{Ablation Analysis}
	We also conduct the ablation analysis to investigate the individual contribution of each constraint. We use each constraint separately to train the network. The visual consistency, semantic dependency, and user-error sparsity achieve 0.572, 0.564 and 0.560 mAP respectively on Mirflickr dataset for the image annotation experiment, while achieving 0.623, 0.615 and 0.601 mAP respectively for the tag refinement experiment. Meanwhile, they achieve 0.386, 0.380 and 0.375 mAP respectively on the NUS-WIDE dataset for the image annotation experiment, while achieving 0.428, 0.425 and 0.421 mAP respectively for the tag refinement experiment. As we can see, each constraint outperforms all the baseline models, and the most improvement comes from the visual consistency. Since the different constraints are designed from the different aspects, combining them can mutually remedy each other, which further improves the overall performance.
	
	\subsection{Visualize the Dynamic Neighbors}
\begin{figure}[t]
		\centering
		\includegraphics[width=0.95\linewidth]{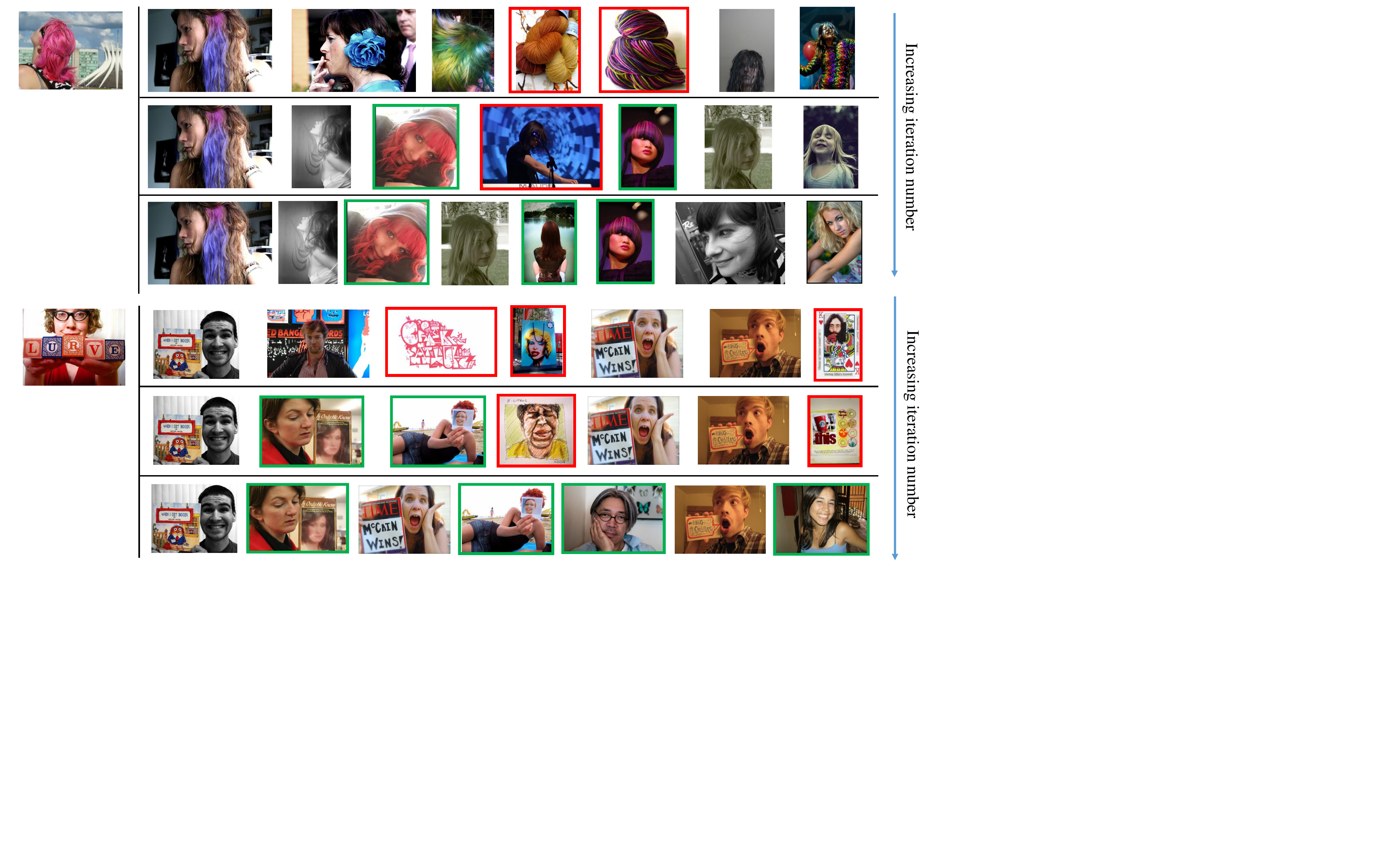}
		\caption{The visualization of the dynamic neighbor selection. The most left image is the query image. Better viewed in color.
		}
		\label{neighbors}
	\end{figure}

	To better demonstrate the effectiveness of our dynamic neighbor selection, given a query image, we show the image neighbors from different iterations in Fig.\ref{neighbors}. As we can see, with the number of iterations increases, the irrelevant neighbors (red boxes) are gradually replaced by the relevant images (green boxes), which proves the effectiveness of our feature learning and dynamic neighbor selection process.
	
	\section{Conclusion}
	The social image annotation and tag refinement have been the research focus since the image sharing networks become popular. In this paper, we propose to solve these problems in a weakly-supervised manner. By dynamically choosing the image neighbors to generate input batches, and formulating the visual consistency, semantic dependency and user-error sparsity as the constraints of the neural network, we can train the proposed model in an end-to-end fashion. Experimental results on two benchmark datasets show that our model outperforms the most methods. Moreover, since the training procedure is similar to a regular CNN, our model is flexible and stable to apply to the large-scale datasets.
	
	{
		\bibliographystyle{aaai}
		\bibliography{egbib}

\begin{thebibliography}{}

\bibitem[\protect\citeauthoryear{Ballan \bgroup et al\mbox.\egroup
  }{2014}]{ballan2014cross}
Ballan, L.; Uricchio, T.; Seidenari, L.; and Del~Bimbo, A.
\newblock 2014.
\newblock A cross-media model for automatic image annotation.
\newblock In {\em {Proc. ACM Int. Conf. Multimedia Retrieval.}}, ~73.
\newblock ACM.

\bibitem[\protect\citeauthoryear{Blei, Ng, and Jordan}{2003}]{blei2003latent}
Blei, D.~M.; Ng, A.~Y.; and Jordan, M.~I.
\newblock 2003.
\newblock Latent dirichlet allocation.
\newblock {\em Journal of machine Learning research} 3(Jan):993--1022.

\bibitem[\protect\citeauthoryear{Bottou}{2010}]{bottou2010large}
Bottou, L.
\newblock 2010.
\newblock Large-scale machine learning with stochastic gradient descent.
\newblock In {\em Proceedings of COMPSTAT'2010}. Springer.
\newblock  177--186.

\bibitem[\protect\citeauthoryear{Chen \bgroup et al\mbox.\egroup
  }{2012}]{chen2012tag}
Chen, L.; Xu, D.; Tsang, I.~W.; and Luo, J.
\newblock 2012.
\newblock Tag-based image retrieval improved by augmented features and
  group-based refinement.
\newblock {\em {{IEEE} Trans. Multimedia}} 14(4):1057--1067.

\bibitem[\protect\citeauthoryear{Chua \bgroup et al\mbox.\egroup
  }{2009}]{chua2009nus}
Chua, T.-S.; Tang, J.; Hong, R.; Li, H.; Luo, Z.; and Zheng, Y.
\newblock 2009.
\newblock Nus-wide: a real-world web image database from national university of
  singapore.
\newblock In {\em Pro. ACM Int. Conf. Image and Video Retrieval}, ~48.
\newblock ACM.

\bibitem[\protect\citeauthoryear{Cilibrasi and
  Vit{\'{a}}nyi}{2007}]{cilibrasi2007google}
Cilibrasi, R., and Vit{\'{a}}nyi, P. M.~B.
\newblock 2007.
\newblock The google similarity distance.
\newblock {\em {IEEE} Trans. Knowl. Data Eng.} 19(3):370--383.

\bibitem[\protect\citeauthoryear{Gong \bgroup et al\mbox.\egroup
  }{2013}]{gong2013deep}
Gong, Y.; Jia, Y.; Leung, T.; Toshev, A.; and Ioffe, S.
\newblock 2013.
\newblock Deep convolutional ranking for multilabel image annotation.
\newblock {\em CoRR} abs/1312.4894.

\bibitem[\protect\citeauthoryear{Guillaumin \bgroup et al\mbox.\egroup
  }{2009}]{guillaumin2009tagprop}
Guillaumin, M.; Mensink, T.; Verbeek, J.; and Schmid, C.
\newblock 2009.
\newblock Tagprop: Discriminative metric learning in nearest neighbor models
  for image auto-annotation.
\newblock In {\em {Proc. IEEE Int. Conf. Comp. Vis.}},  309--316.
\newblock IEEE.

\bibitem[\protect\citeauthoryear{Huiskes and Lew}{2008}]{huiskes2008mir}
Huiskes, M.~J., and Lew, M.~S.
\newblock 2008.
\newblock The mir flickr retrieval evaluation.
\newblock In {\em Pro. ACM Int. Conf. Multimedia Info. Retrieval},  39--43.
\newblock ACM.

\bibitem[\protect\citeauthoryear{Krizhevsky, Sutskever, and
  Hinton}{2012}]{krizhevsky2012imagenet}
Krizhevsky, A.; Sutskever, I.; and Hinton, G.~E.
\newblock 2012.
\newblock Imagenet classification with deep convolutional neural networks.
\newblock In {\em {Proc. Advances in Neural Inf. Process. Syst.}},  1097--1105.

\bibitem[\protect\citeauthoryear{Li and Snoek}{2013}]{li2013classifying}
Li, X., and Snoek, C.~G.
\newblock 2013.
\newblock Classifying tag relevance with relevant positive and negative
  examples.
\newblock In {\em {Proc. {ACM} Int. Conf. Multimedia.}},  485--488.
\newblock ACM.

\bibitem[\protect\citeauthoryear{Li and Tang}{2017a}]{li2017weaklymatrix}
Li, Z., and Tang, J.
\newblock 2017a.
\newblock Weakly supervised deep matrix factorization for social image
  understanding.
\newblock {\em {{IEEE} Trans. Image Process.}} 26(1):276--288.

\bibitem[\protect\citeauthoryear{Li and Tang}{2017b}]{li2017weakly}
Li, Z., and Tang, J.
\newblock 2017b.
\newblock Weakly-supervised deep nonnegative low-rank model for social image
  tag refinement and assignment.
\newblock In {\em {Proc. Conf. AAAI}},  4154--4160.

\bibitem[\protect\citeauthoryear{Li \bgroup et al\mbox.\egroup
  }{2016}]{li2016socializing}
Li, X.; Uricchio, T.; Ballan, L.; Bertini, M.; Snoek, C.~G.; and Bimbo, A.~D.
\newblock 2016.
\newblock Socializing the semantic gap: A comparative survey on image tag
  assignment, refinement, and retrieval.
\newblock {\em ACM Computing Surveys} 49(1):14.

\bibitem[\protect\citeauthoryear{Li, Snoek, and Worring}{2009}]{li2009learning}
Li, X.; Snoek, C.~G.; and Worring, M.
\newblock 2009.
\newblock Learning social tag relevance by neighbor voting.
\newblock {\em {{IEEE} Trans. Multimedia}} 11(7):1310--1322.

\bibitem[\protect\citeauthoryear{Liu \bgroup et al\mbox.\egroup
  }{2009}]{liu2009tag}
Liu, D.; Hua, X.-S.; Yang, L.; Wang, M.; and Zhang, H.-J.
\newblock 2009.
\newblock Tag ranking.
\newblock In {\em {Proc. Int. Conf. World Wide Web.}},  351--360.
\newblock ACM.

\bibitem[\protect\citeauthoryear{Makadia, Pavlovic, and
  Kumar}{2010}]{makadia2010baselines}
Makadia, A.; Pavlovic, V.; and Kumar, S.
\newblock 2010.
\newblock Baselines for image annotation.
\newblock {\em {Int. J. Comput. Vision}} 90(1):88--105.

\bibitem[\protect\citeauthoryear{Miller}{1995}]{miller1995wordnet}
Miller, G.~A.
\newblock 1995.
\newblock Wordnet: a lexical database for english.
\newblock {\em Communications of the ACM} 38(11):39--41.

\bibitem[\protect\citeauthoryear{Murthy, Maji, and
  Manmatha}{2015}]{murthy2015automatic}
Murthy, V.~N.; Maji, S.; and Manmatha, R.
\newblock 2015.
\newblock Automatic image annotation using deep learning representations.
\newblock In {\em {Proc. ACM Int. Conf. Multimedia Retrieval.}},  603--606.
\newblock ACM.

\bibitem[\protect\citeauthoryear{Pereira \bgroup et al\mbox.\egroup
  }{2014}]{pereira2014role}
Pereira, J.~C.; Coviello, E.; Doyle, G.; Rasiwasia, N.; Lanckriet, G.~R.; Levy,
  R.; and Vasconcelos, N.
\newblock 2014.
\newblock On the role of correlation and abstraction in cross-modal multimedia
  retrieval.
\newblock {\em {{IEEE} Trans. Pattern Anal. Mach. Intell.}} 36(3):521--535.

\bibitem[\protect\citeauthoryear{Sigurbj{\"o}rnsson and
  Van~Zwol}{2008}]{sigurbjornsson2008flickr}
Sigurbj{\"o}rnsson, B., and Van~Zwol, R.
\newblock 2008.
\newblock Flickr tag recommendation based on collective knowledge.
\newblock In {\em {Proc. Int. Conf. World Wide Web.}},  327--336.
\newblock ACM.

\bibitem[\protect\citeauthoryear{Simonyan and
  Zisserman}{2014}]{simonyan2014very}
Simonyan, K., and Zisserman, A.
\newblock 2014.
\newblock Very deep convolutional networks for large-scale image recognition.
\newblock {\em CoRR} abs/1409.1556.

\bibitem[\protect\citeauthoryear{Wan \bgroup et al\mbox.\egroup
  }{2014}]{wan2014deep}
Wan, J.; Wang, D.; Hoi, S. C.~H.; Wu, P.; Zhu, J.; Zhang, Y.; and Li, J.
\newblock 2014.
\newblock Deep learning for content-based image retrieval: A comprehensive
  study.
\newblock In {\em {Proc. {ACM} Int. Conf. Multimedia.}},  157--166.
\newblock ACM.

\bibitem[\protect\citeauthoryear{Wang \bgroup et al\mbox.\egroup
  }{2016}]{wang2016cnn}
Wang, J.; Yang, Y.; Mao, J.; Huang, Z.; Huang, C.; and Xu, W.
\newblock 2016.
\newblock {CNN-RNN:} {A} unified framework for multi-label image
  classification.
\newblock {\em {Proc. IEEE Conf. Comp. Vis. Patt. Recogn.}}  2285--2294.

\bibitem[\protect\citeauthoryear{Wei \bgroup et al\mbox.\egroup
  }{2016}]{single2multi}
Wei, Y.; Xia, W.; Lin, M.; Huang, J.; Ni, B.; Dong, J.; Zhao, Y.; and Yan, S.
\newblock 2016.
\newblock Hcp: A flexible cnn framework for multi-label image classification.
\newblock {\em {{IEEE} Trans. Pattern Anal. Mach. Intell.}} 38(9):1901--1907.

\bibitem[\protect\citeauthoryear{Xu \bgroup et al\mbox.\egroup
  }{2009}]{xu2009tag}
Xu, H.; Wang, J.; Hua, X.-S.; and Li, S.
\newblock 2009.
\newblock Tag refinement by regularized lda.
\newblock In {\em {Proc. {ACM} Int. Conf. Multimedia.}},  573--576.
\newblock ACM.

\bibitem[\protect\citeauthoryear{Zhu, Ngo, and Jiang}{2012}]{zhu2012sampling}
Zhu, S.; Ngo, C.-W.; and Jiang, Y.-G.
\newblock 2012.
\newblock Sampling and ontologically pooling web images for visual concept
  learning.
\newblock {\em {{IEEE} Trans. Multimedia}} 14(4):1068--1078.

\bibitem[\protect\citeauthoryear{Zhu, Yan, and Ma}{2010}]{zhu2010image}
Zhu, G.; Yan, S.; and Ma, Y.
\newblock 2010.
\newblock Image tag refinement towards low-rank, content-tag prior and error
  sparsity.
\newblock In {\em {Proc. {ACM} Int. Conf. Multimedia.}},  461--470.
\newblock ACM.

\end{thebibliography}
	}
	
\end{document}